\documentclass[pmlr]{jmlr} 

\newcommand{\tagdsmode}{proceedings}

\makeatletter
\newcommand{\tagdssubmission}{submission}
\newcommand{\tagdsproceedings}{proceedings}

\ifx\tagdsmode\tagdsproceedings

\else\ifx\tagdsmode\tagdssubmission
  \def\ps@jmlrtps{%
    \let\@mkboth\@gobbletwo
    \def\@oddhead{\scriptsize Under Review at the 2nd Conference on Topology, Algebra, and Geometry in Data Science\hfill}%
    \let\@evenhead\@oddhead
    \def\@oddfoot{}%
    \let\@evenfoot\@oddfoot
  }

\else
  \def\ps@jmlrtps{%
    \let\@mkboth\@gobbletwo
    \def\@oddhead{}%
    \let\@evenhead\@oddhead
    \def\@oddfoot{}%
    \let\@evenfoot\@oddfoot
  }
\fi\fi
\makeatother

\usepackage{longtable}

\usepackage{booktabs}
\usepackage[load-configurations=version-1]{siunitx} 

\theorembodyfont{\upshape}
\theoremheaderfont{\scshape}
\theorempostheader{:}
\theoremsep{\newline}

\jmlrvolume{334}
\jmlryear{2026}
\jmlrworkshop{Topology, Algebra, and Geometry in Data Science}

\title[Topology in PCNs]{Topological Simplification in Predictive Coding Networks}

\ifx\tagdsmode\tagdssubmission

\else

 \author{\Name{Adam Shaw} \Email{adshaw@usc.edu}\\
  \Name{Jiayu Li} \Email{jli99757@usc.edu}\\
  \Name{Michael Sperling} \Email{mesperli@usc.edu}\\
  \Name{Michael Kim} \Email{mkim2763@usc.edu}\\
  \Name{Alvin Jin} \Email{alvinjin@usc.edu}\\
  \addr University of Southern California \\ 3551 Trousdale Parkway, Los Angeles, CA 90089}

\fi

\begin{document}

\maketitle

\begin{abstract}
We study the topology of learned representations in predictive coding networks (PCNs), a neuro-inspired bidirectional architecture, using a quantitative layer-wise persistent homology analysis. We train well-performing PCNs on a synthetic classification dataset ($\geq 99.9\%$ test accuracy) and on MNIST ($\geq 95\%$ test accuracy), and measure how topological features change across layers for different architectures and activation functions. We find that smaller PCNs collapse connected components across layers earlier than larger models (Spearman $\rho \in [0.72, 0.79]$ across activations), with model size measured as the sum of hidden-layer widths. We also observe a strong negative correlation ($\rho = -0.58$) between the depth at which simplification occurs and reconstruction error; i.e., architectures that simplify later reconstruct better. Finally, a seed-level bootstrap comparison across architectures and activations shows that PCNs consistently collapse connected components later than matched MLPs, with an average difference of $3.6$ layers. These results suggest that persistent homology offers a useful quantitative lens on the compression--reconstruction tradeoff in PCNs, and that both model capacity and the recurrent, bidirectional dynamics of predictive coding inference shape when this tradeoff is resolved across layers.
\end{abstract}


\section{Introduction}
\label{sec:intro}
Several studies have used persistent homology \citep{edelsbrunner2002persistence} to analyze how topological structure evolves within deep neural network representations, showing that feedforward networks progressively simplify topology by collapsing connected components, loops, and other homological features \citep{Naitzat2020, Watanabe_Yamana_2021, suresh2024ph, ergen2024relunets}. Whether this phenomenon extends beyond feedforward architectures remains unclear. In particular, little is known about how topology evolves in models that must simultaneously support inference and reconstruction, such as predictive coding networks (PCNs) \citep{rao1999predictive}.


In this work, we use persistent homology to study the evolution of representation topology in PCNs trained on both synthetic and real-world datasets. To this end, we introduce two metrics: one that quantifies when topological simplification occurs across depth, and one that measures reconstruction and inversion quality.

Our results show that PCNs exhibit progressive topological simplification while keeping some ability for inversion. Smaller models simplify topology earlier than larger models, and earlier simplification is associated with poorer reconstruction performance. Together, these findings extend topological analyses of learned representations beyond feedforward networks and highlight a connection between topological simplification and reconstruction in generative architectures.

\section{Background}
\label{sec:background}

\subsection{Predictive Coding Networks}
\label{sec:pcn_intro}
Predictive coding networks (PCNs) are hierarchical generative models that perform inference through iterative minimization of a global energy functional rather than a single feedforward computation \citep{rao1999predictive, whittington2017approximation, Stenlund2025}. Unlike conventional feedforward architectures, PCNs support approximate inversion of learned representations through the same recurrent dynamics used for inference \citep{rao1999predictive, sun2020inversion}. From a topological perspective, this makes PCNs a particularly informative test case: whereas feedforward classifiers may freely collapse topological structure to produce separable representations, predictive coding networks must preserve sufficient information to support reconstruction.




PCNs have also attracted interest as a biologically-plausible alternative to backpropagation \citep{backprop} by avoiding the \textit{weight transport problem} \citep{GROSSBERG198723}: While backpropagation requires error signals to pass through exact transposes of downstream weight matrices (a biologically-implausible operation), PCNs use local interactions between neighboring layers. Despite this locality constraint, predictive coding updates approximate backpropagation to a high degree \citep{whittington2017approximation}. We lay out more details regarding the PCN formulation we use in Appendix~\ref{sec:PCN_appendix}.

\begin{figure}[htbp]
    \centering
    \includegraphics[width=0.91\linewidth]{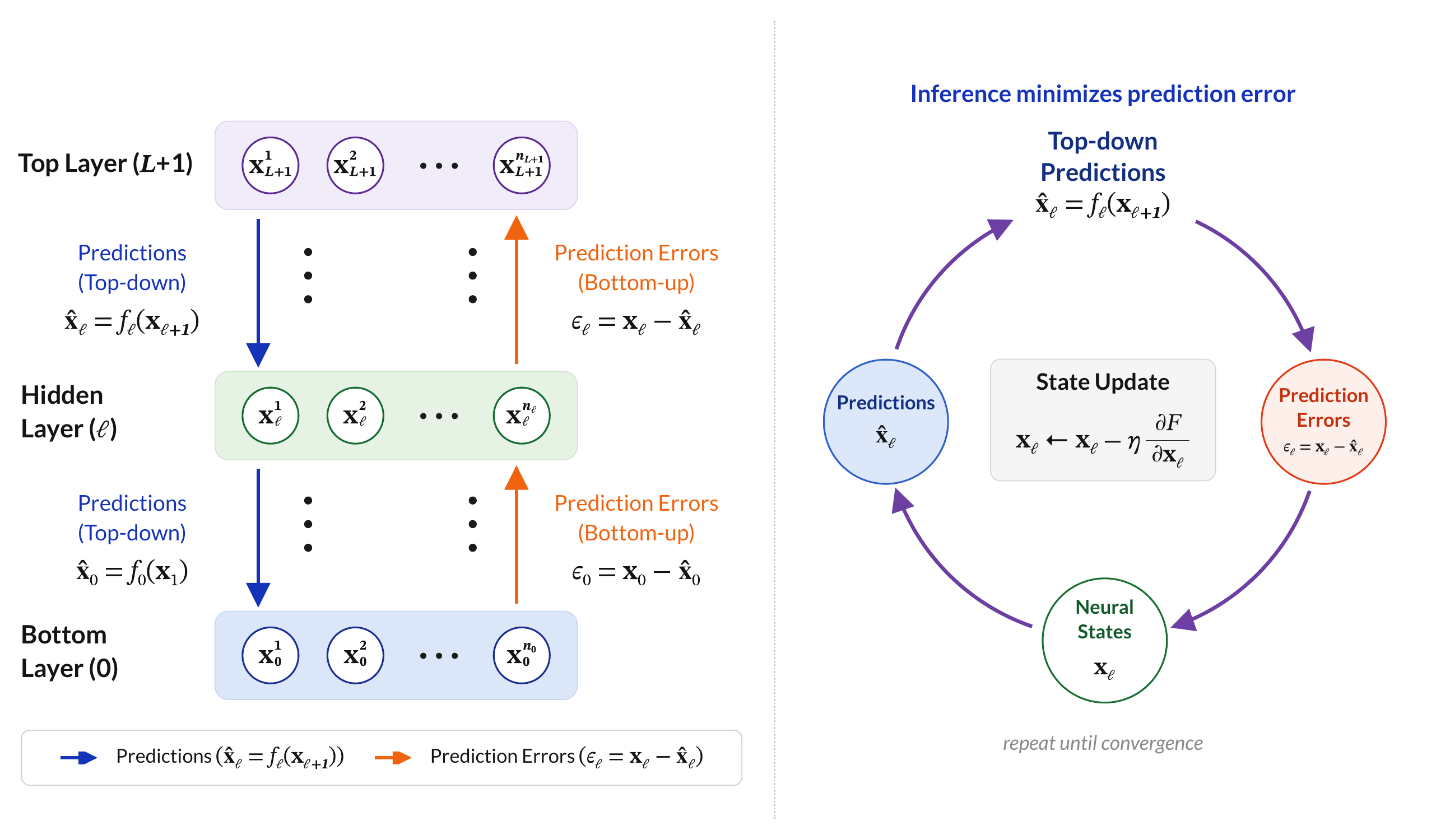}
    \vspace{-\baselineskip}
    \caption{Left: Predictive Coding Network architecture; Right: PCN inference dynamics \citep{whittington2017approximation, rao1999predictive}.}
    \label{fig:PCN}
\end{figure}

\subsection{Topological Simplification in Neural Network Representations}
\label{sec:ph}

Persistent homology provides a framework for quantifying topological structure in finite point clouds by summarizing homological features across scales as a persistence barcode \citep{edelsbrunner2002persistence}. Persistent homology distinguishes robust topological structure from sampling noise by identifying features that persist across scales, making it particularly well-suited for analyzing high-dimensional neural representations inferred from finite samples. In this work, we use persistent homology as a descriptive tool for tracking the evolution of topological complexity across layers of predictive coding networks.

\section{Methodology}
\label{sec:methodology}

\subsection{Objective}
\label{sec:objective}
Our experiments test whether PCNs exhibit topological simplification and identify the architectural factors that govern this behavior. We analyze how topological invariants evolve across layers in PCNs, comparing their behavior to standard feedforward networks to assess whether topological simplification may be a general property of deep representations or depends on specific architectural choices. To isolate these effects, we systematically vary hidden-layer width and activation function, measuring how these choices influence the timing and extent of Betti number decay. We further examine how topological simplification relates to the approximate invertibility enabled by predictive coding dynamics, testing whether pressure to support reconstruction constrains excessive topological collapse. All models are implemented using the \texttt{JAX}-based \texttt{PCX} library \citep{jax, pinchetti2025pcx}.

\subsection{Datasets}
\label{sec:datasets}
We conduct our experiments on both synthetic and real-world datasets. For synthetic data, we sample from a topologically-complex, two-dimensional manifold made up of $M_a$ (nine disjoint disks, i.e., with $\beta_0 = 9,\ \beta_1 = 0$) positioned within $M_b$ (a single connected region containing nine holes, i.e., with $\beta_0 = 1,\ \beta_1 = 9$); see \figureref{fig:D1_picture}. This dataset follows the construction introduced by \citet{Naitzat2020}, which we adopt because the classes are not linearly separable or even separable by any simple topological deformation, meaning any classifier must perform significant topological transformations on the data to pull the classes apart in feature space. To complement this analysis, we also evaluate our approach on MNIST \citep{lecun1998mnist}, a widely used real-world image dataset whose intrinsic topology is unknown and substantially more complex.

\begin{figure}[htbp]
\floatconts
    {fig:D1_picture}
    {
        \caption{Visualization of the synthetic dataset, consisting of manifold $M_a$ (green) embedded in manifold $M_b$ (red), adapted from \cite{Naitzat2020}.}
        \vspace{-\baselineskip}
    }
    {
        \includegraphics[width=0.3\linewidth]{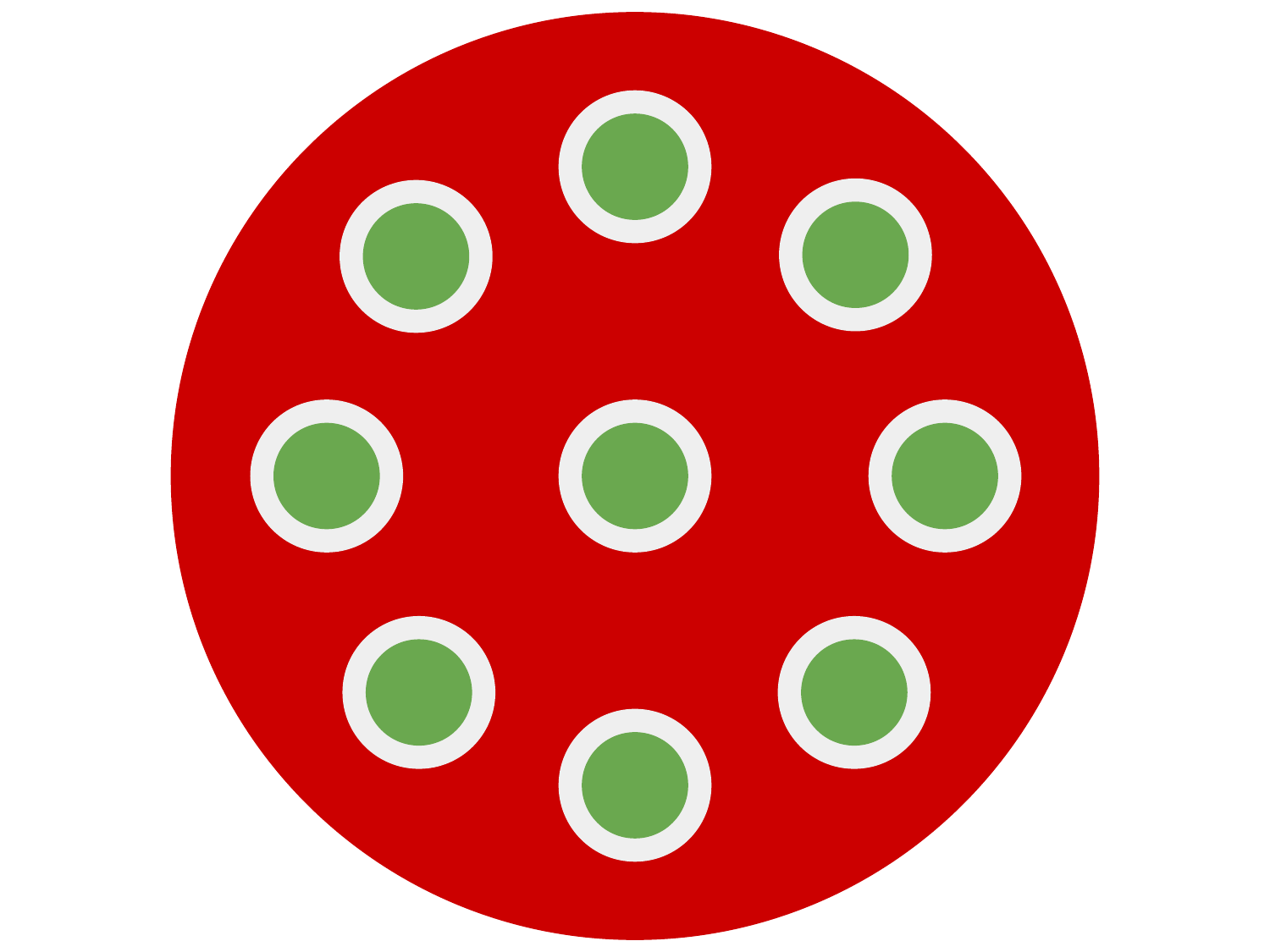}
        \vspace{-\baselineskip}
    }
\end{figure}

\subsection{Model Variants}
For both the synthetic dataset and MNIST, we evaluate a range of architectures that vary in hidden-layer width and in how that width progresses across layers, including uniform-width networks, networks with an increasing/decreasing width, and networks with a bottleneck in the middle. Despite this variety, all models share a fixed depth of eight hidden layers, so that topological simplification at each layer can be directly compared across architectures. For each dataset, we train 30 independent instances of each model configured using distinct random seeds. Synthetic data models were trained to at least $99.9\%$ test accuracy, while models trained on MNIST achieved at least $95\%$ test accuracy due to increased complexity.

Throughout this paper, we refer to a model as having $L$ hidden layers, with layer indices $\ell \in \{0, \ldots, L+1\}$ including the input and output layers. A complete list of all architectures used is provided in \appendixref{app:model_details}.

\subsection{Topology Tracking Procedure}
\label{sec:topology_tracking}

We adopt and extend the topological analysis framework of \citet{Naitzat2020}, with help from \citet{wheeler2021}, to quantify how networks simplify representation topology across depth. For each layer $\ell$, we collect the hidden representations of a fixed set of data points and construct a metric space using either a normalized $k$-nearest neighbors ($k$-NN) graph with unweighted geodesic distance (for synthetic data) or a normalized Euclidean distance (for real-world data), both of which mitigate arbitrary scaling differences across layers. Persistent homology is then computed on this metric space, and Betti numbers are evaluated at a fixed filtration scale $\eta$ to obtain per-layer scalar summaries. To avoid conflating the topologies of different classes, we restrict persistent homology computations to a single class per dataset: manifold $M_a$ for the synthetic data, and the digit-$0$ class for MNIST. Formal implementation details (including the choice of $k$ and $\eta$) are provided in \appendixref{sec:topology_tracking_appendix}.

\subsection{Metrics}
\label{sec:metrics}

We define two new metrics: one to quantify timing of topological simplification, and another to quantify reconstruction fidelity.

\subsubsection{Center of Mass of Topological Simplification}
\label{sec:com}

Let $B(\ell) = \sum_{k \in \mathcal{K}} w_k \beta_k^{(\ell)}$ be a weighted sum of Betti numbers at layer $\ell$, for some range of homological degrees (e.g., $\mathcal{K} = \{0, 1\}$ for $\beta_0$ and $\beta_1$), and let $\tilde{B}(\ell) = \min_{j \leq \ell} B(j)$ be its running minimum, so that simplification events are treated as irreversible. The layer-wise Betti drop $\Delta(\ell) = \tilde{B}(\ell-1) - \tilde{B}(\ell)$ then defines a distribution over layers $p(\ell) = \Delta(\ell) / D$, where $D = \sum_\ell \Delta(\ell)$ is the total simplification across all layers. If $D=0$, the distribution $p(\ell)$ is undefined and $\mathrm{COM}$ is left unassigned; this case did not arise in any of our experiments.

\begin{definition}[COM]
\label{def:com}
To quantify when simplification occurs, we define the \emph{Center of Mass of Topological Simplification (COM)} as the expected depth of topological simplification under this distribution:
\begin{equation}
\label{eq:com_drop}
\mathrm{COM}
\;=\;
\sum_{\ell=1}^{L+1} \ell \, p(\ell)
\;=\;
\frac{\sum_{\ell=1}^{L+1} \ell \, \Delta(\ell)}{\sum_{\ell=1}^{L+1} \Delta(\ell)}.
\end{equation}
\end{definition}

A lower COM indicates earlier topological collapse. For conciseness, we write $\mathrm{COM}(\beta_0 + \beta_1)$ as shorthand to indicate that $B(\ell) = \beta_0^{(\ell)} + \beta_1^{(\ell)}$, and similarly for other linear combinations of Betti numbers.

\subsubsection{Mean Reconstruction Distance}
\label{sec:mrd}

To quantify reconstruction fidelity, we measure how closely inverted outputs resemble valid same-class inputs. For each of $M=30$ independently-trained models $m$ of an architecture $A$, we pass $N=1000$ synthetic one-hot outputs $y_i$ through the inversion procedure (see Appendix~\ref{sec:inversion_appendix}), each yielding a reconstruction $\hat{x}_i^{(m)}$. We then find its nearest Euclidean neighbor $x_i^{*(m)}$ in the training set restricted to the same class.

\begin{definition}[MRD]
\label{def:mrd}
We define the \emph{Mean Reconstruction Distance (MRD)} of an architecture $A$ to be the mean Euclidean distance between reconstructions and their nearest same-class neighbors, averaged over $N$ reconstructions and $M$ model seeds:
\begin{equation}
    \mathrm{MRD}(A) = \frac{1}{MN} \sum_{m=1}^{M} \sum_{i=1}^N \left\|\hat{x}_i^{(m)} - x_i^{*(m)}\right\|_2.
\end{equation}
\end{definition}
By comparing to the nearest same-class neighbor rather than a fixed target, MRD captures whether inversions lie near the correct data manifold without penalizing valid novel reconstructions.

See Figure~\ref{fig:recons_comparison} for a comparison of the inversion behavior of two models with substantially different MRD values.

\section{Results}
\label{sec:results}
Across PCN architectures, we observe a consistent trend of topological simplification, with the timing and extent of this simplification varying systematically with architectural capacity and activation function, and correlating with reconstruction quality. This analysis was conducted through a multitude of experiments available publicly on GitHub.\footnote{https://github.com/jiayuliusc/Topological-Simplification-in-Predictive-Coding-Networks}

\begin{figure}[htbp]
\floatconts
    {fig:betti_trends}
    {
        \caption{Zeroth Betti numbers ($\beta_0$) across layers for two eight-layer ReLU architectures. Each plot shows the mean $\beta_0$ at each layer, its $\pm 1$ SD band, and individual seeds. The monotonic decline in the alternating-width model (right) suggests genuine topological simplification rather than simple geometric compression and re-expansion between layers.}
    }
    {%
        \vspace{0.5\baselineskip}
        \subfigure[$\beta_0$ for 30x4 + 18x4 ReLU] {%
            \label{fig:uniform_relu_b0}
            \includegraphics[width=0.44\linewidth]{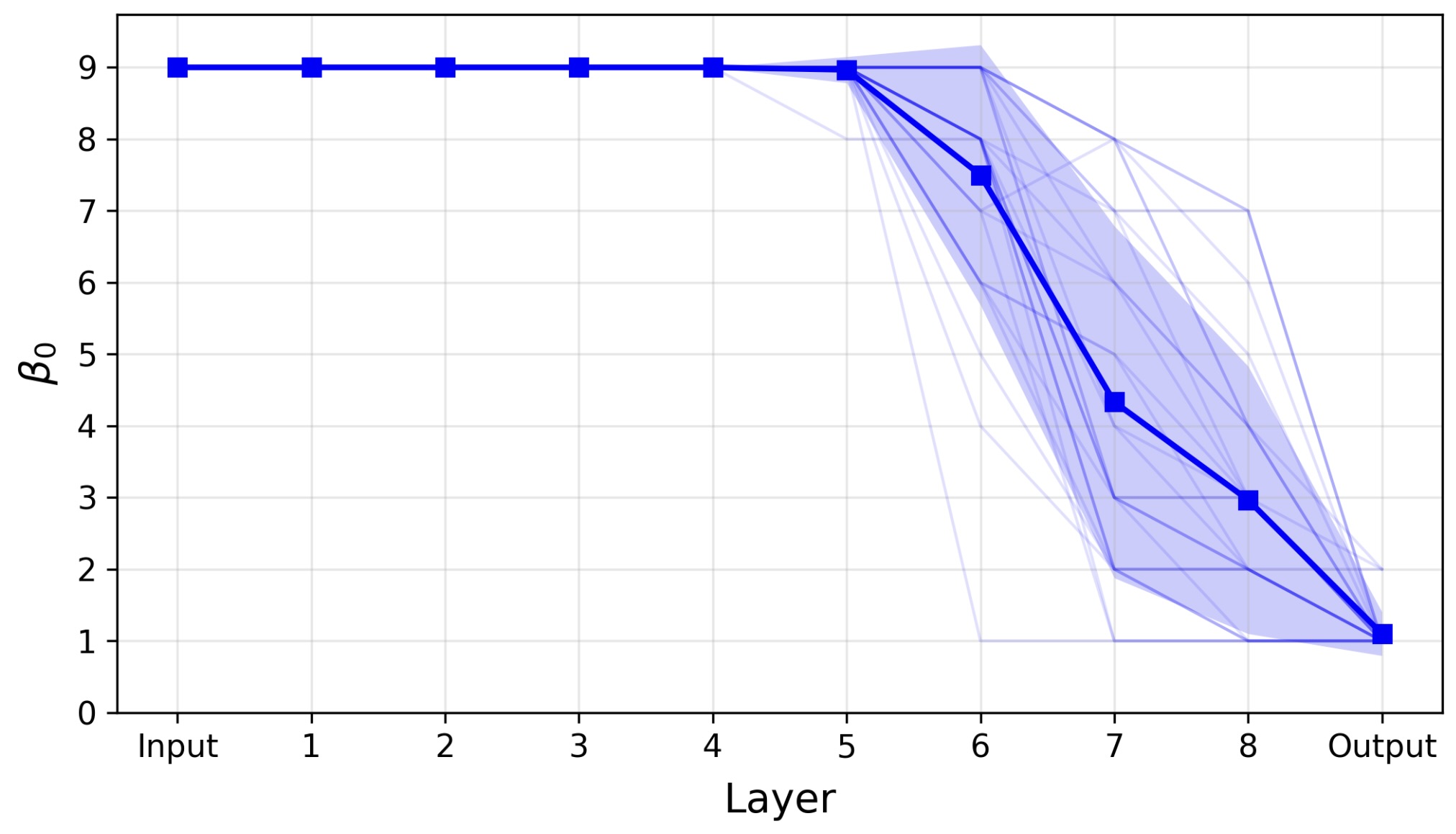}
        }
        \subfigure[$\beta_0$ for Alternating (30,\! 15) ReLU] {%
            \label{fig:alt_relu_b0}
            \includegraphics[width=0.44\linewidth]{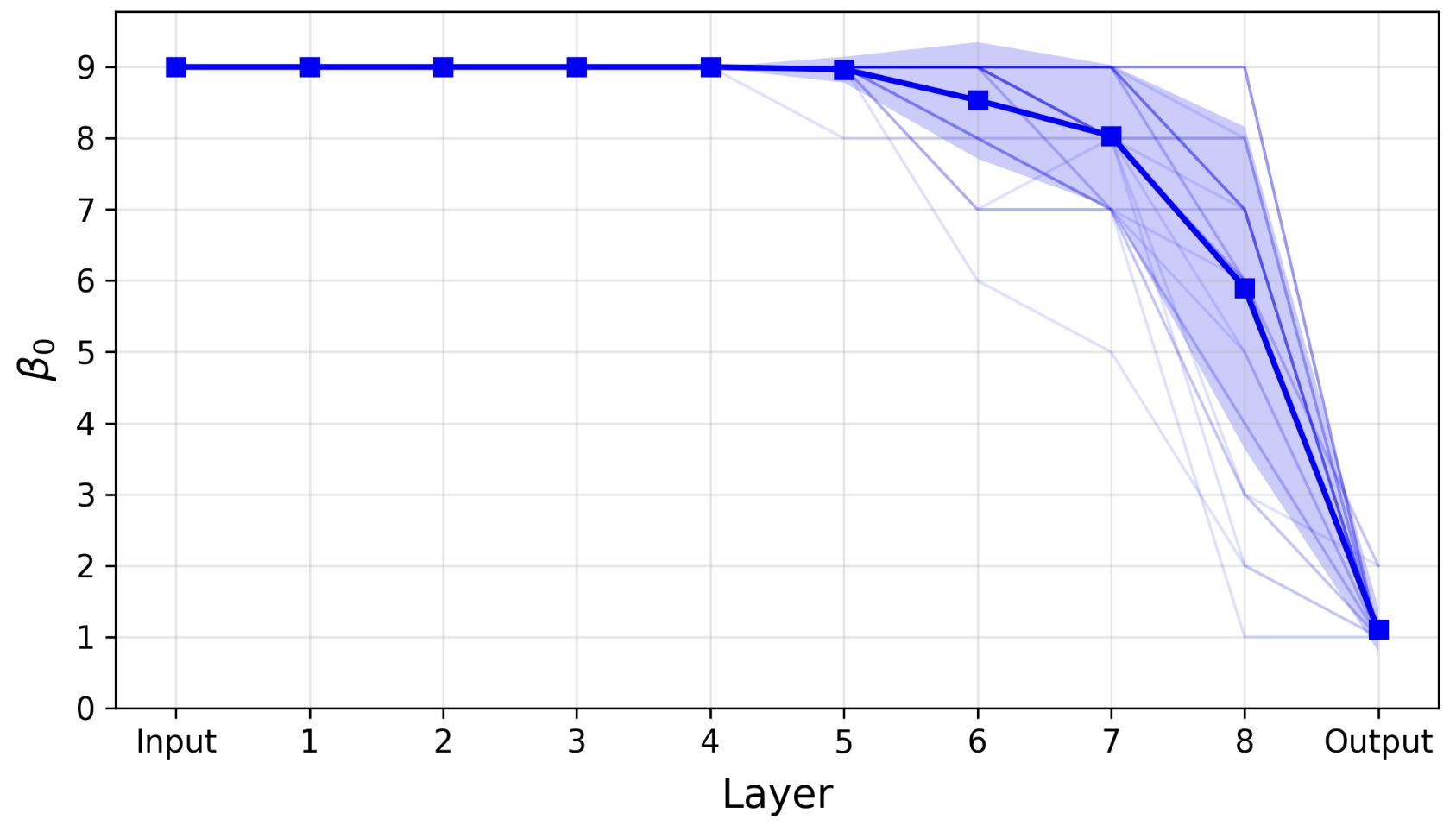}
        }
        \vspace{-\baselineskip}
    }
\end{figure}

\subsection{Synthetic Data}

\paragraph{Architectural Constraints}

Our results indicate that smaller models exhibit topological simplification earlier, on average, than larger models. To quantify this relationship, we examine the correlation between simplification timing (COM) and model size, measured as the sum of hidden-layer widths. As shown in Figure~\ref{fig:com_vs_hidden}, simplification timing is strongly positively correlated with model size across activation functions, with Spearman correlation coefficients of $0.76$ for ReLU, $0.79$ for Leaky ReLU, and $0.72$ for $\tanh$. This stratification is necessary because the activation function is a confounding variable that influences both representation geometry and simplification behavior. A likely explanation for these results is that larger models have the capacity to apply more complex geometric transformations while preserving the overall shape of the data manifold, whereas smaller models may be forced to collapse this structure earlier in order to produce a separable representation.

\begin{figure}[t]
\floatconts
    {fig:com_vs_hidden}
    {
        \caption{Model size versus simplification timing ($\mathrm{COM}(\beta_0)$) on the synthetic dataset, shown separately for each activation function.}
        \vspace{-0.5\baselineskip}
    }
    {%
        \subfigure[ReLU activations]{%
            \label{fig:all_relu_com_vs_hidden}
            \includegraphics[width=0.31\linewidth]{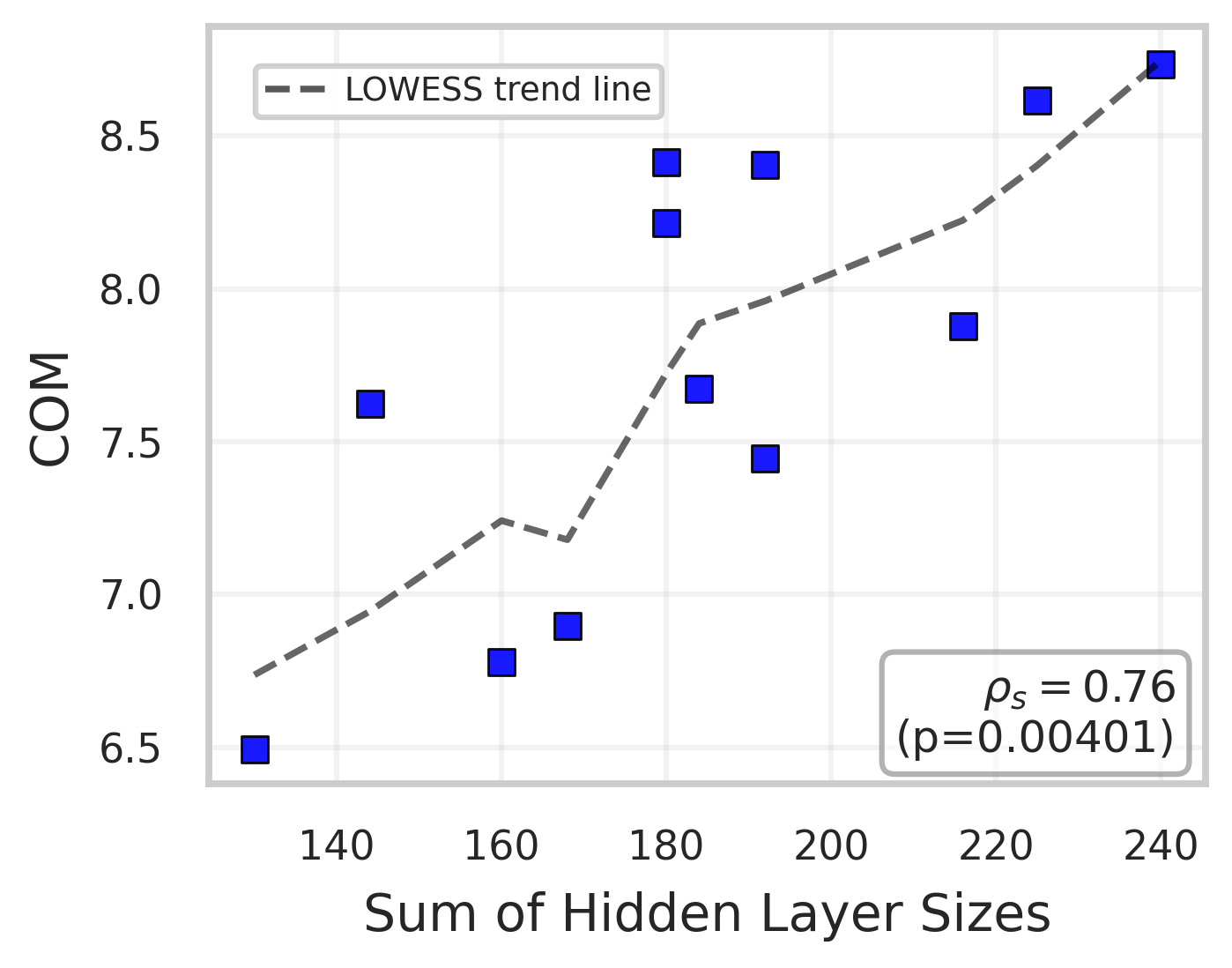}%
        }
        \subfigure[Leaky ReLU activations]{%
            \label{fig:all_leaky_com_vs_hidden}
            \includegraphics[width=0.31\linewidth]{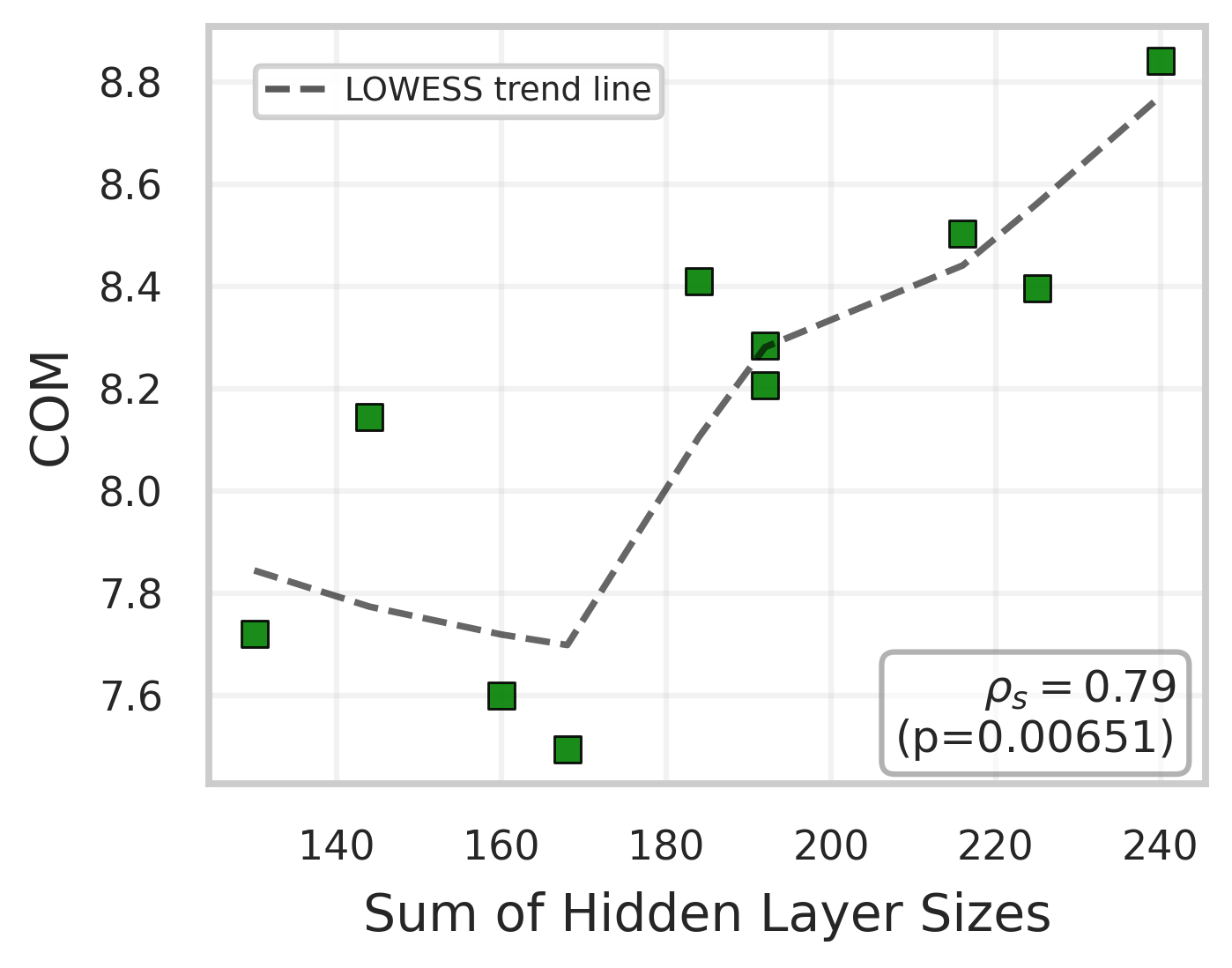}%
        }
        \subfigure[$\tanh$ activations]{%
            \label{fig:all_tanh_com_vs_hidden}
            \includegraphics[width=0.31\linewidth]{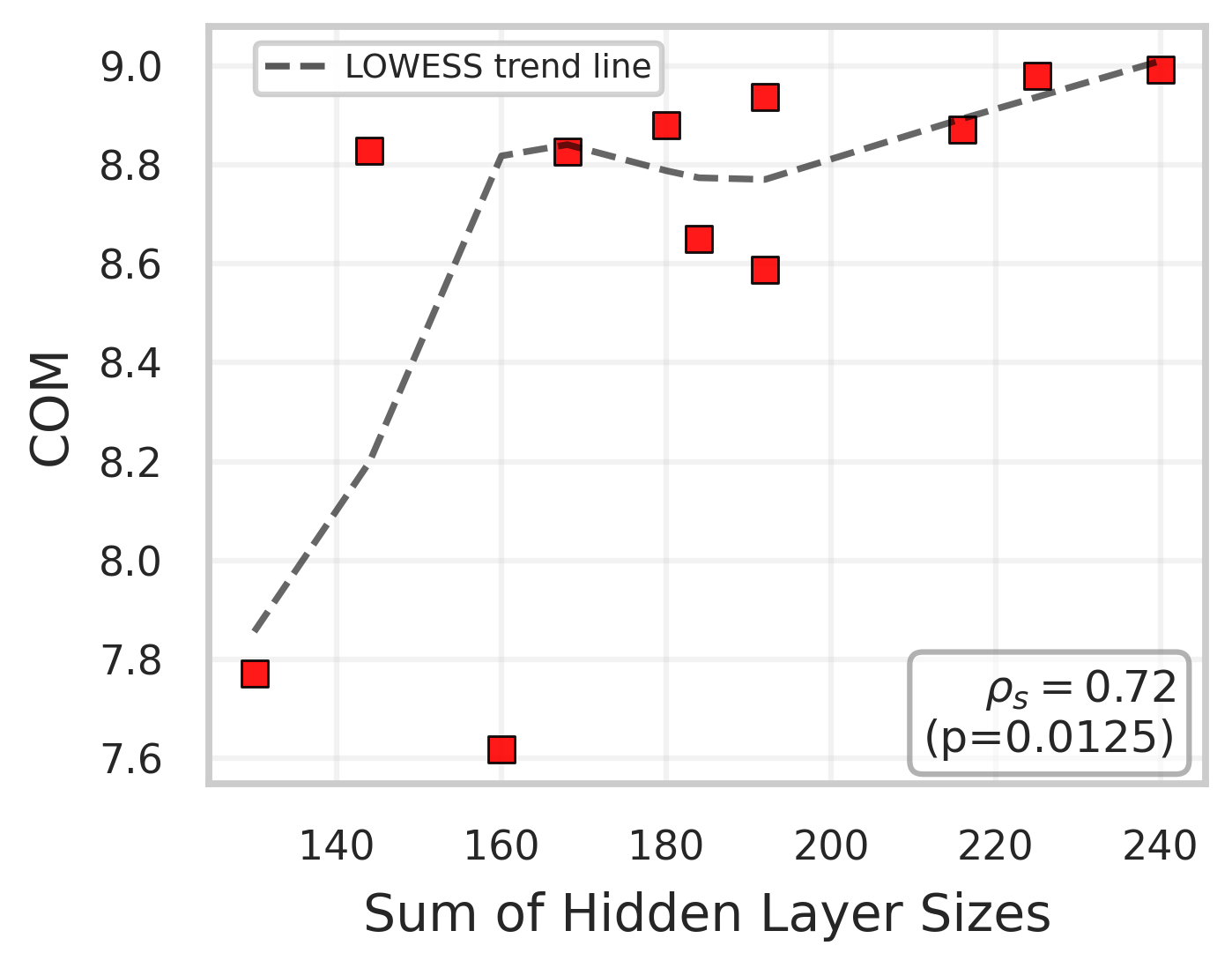}%
        }
        \vspace{-\baselineskip}
    }
\end{figure}

\paragraph{Capacity Sensitivity by Activation Function}
To further investigate this relationship across activation functions, we fit a separate linear model for each activation,
\begin{equation}
    \mathrm{COM}_a(P)=\alpha_a+\gamma_a P+\varepsilon,
\end{equation}
where $P=\sum_{\ell=1}^L n_\ell$ is the total hidden width of the architecture and $a$ denotes the activation function. The fitted slopes were positive for all three activations: $100\hat{\gamma}=1.79$ for ReLU, $0.96$ for Leaky ReLU, and $0.91$ for $\tanh$, with corresponding $R^2$ values of $0.59$, $0.63$, and $0.41$, respectively. Thus, while increasing capacity delays simplification for all activations, the effect is strongest for ReLU: its slope is nearly twice that of Leaky ReLU and $\tanh$. This suggests that ReLU PCNs are more capacity-sensitive, in the sense that added width is associated with a substantially greater delay in topological collapse than in PCNs using other activation functions.

\paragraph{Simplification-Reconstruction Tradeoff}
We observe a strong negative correlation between COM and MRD ($\rho = -0.58, \; p < 10^{-3}$), indicating that architectures that delay topological simplification preserve invertibility; see \figureref{fig:mrd_com}. Intuitively, COM measures when irreversible geometric mergers occur in representation space, while MRD measures how much structure remains available for inversion. Collapsing connected components corresponds to identifying previously distinct regions of the input manifold, so when this occurs early, information is lost, which harms reconstruction quality.

\begin{figure}[htbp]
\floatconts
    {fig:mrd_com}
    {
        \caption{Graphical relationship and correlation between early topological simplification ($\mathrm{COM}(\beta_0)$) and reconstruction error (MRD) in architectures trained on the synthetic dataset. The strong negative correlation highlights a clear association between earlier topological simplification and weaker reconstruction performance.}
        \vspace{-1.5\baselineskip}
    }
    {
        \includegraphics[width=0.7\linewidth]{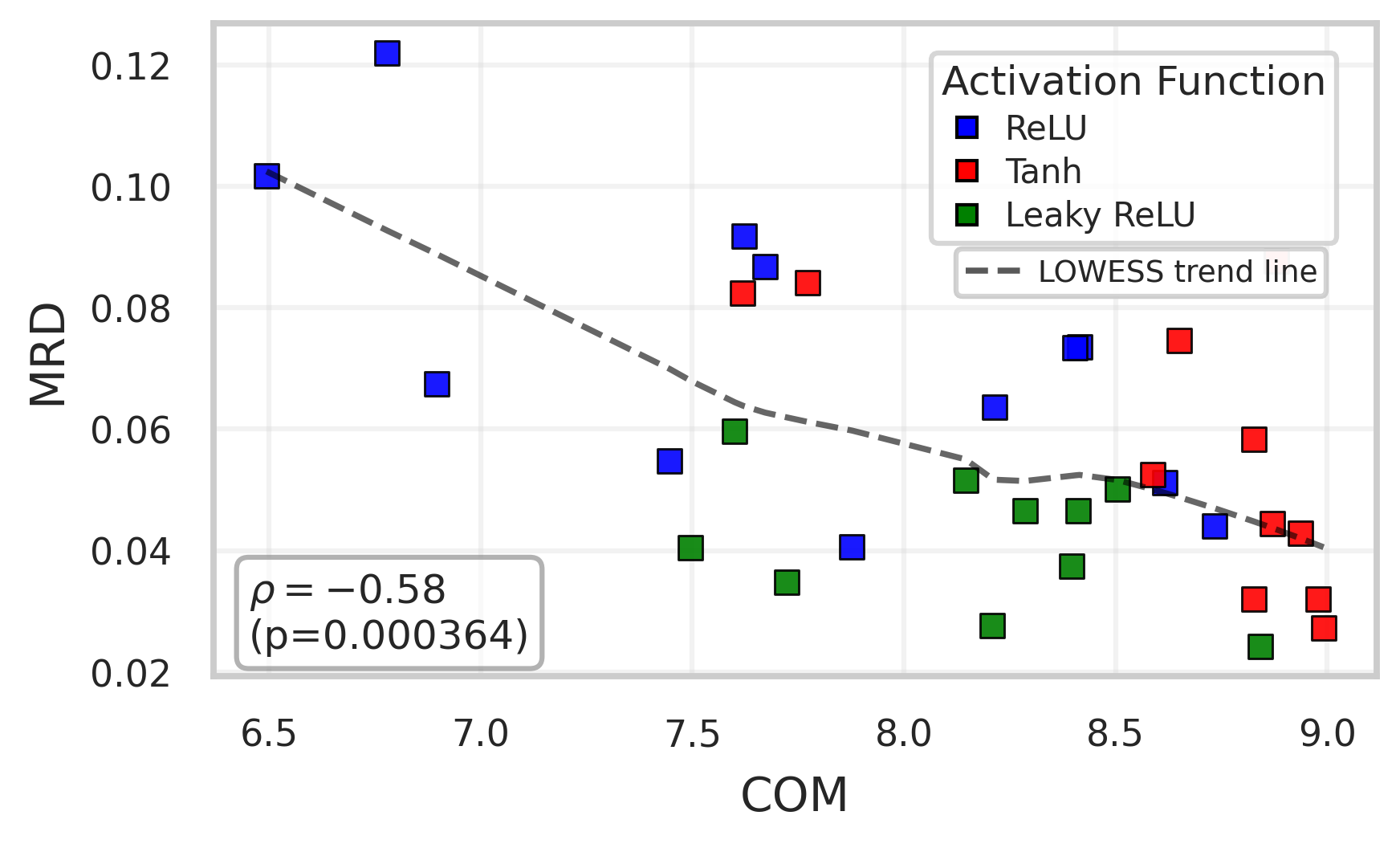}
        \vspace{-1.5\baselineskip}
    }
\end{figure}



\subsection{Consistency with Real-World Data}
\label{sec:mnist_analysis}

We extend our analysis to the MNIST dataset of handwritten digits \citep{lecun1998mnist} to evaluate whether the topological simplification phenomena observed in controlled synthetic settings persist in realistic, high-dimensional data. Due to the substantial computational cost of persistent homology estimation on high-dimensional point clouds, we restrict our MNIST experiments to a set of representative architectures, detailed in \tableref{tab:mnist_models}. Following the manifold view of image data \citep{fefferman2016testing,Naitzat2020}, we treat each digit class as a point cloud in $\mathbb{R}^{784}$ and track $\beta_0$ and $\beta_1$ across layers of trained PCNs. We use a Vietoris–Rips scale parameter across $\eta \in [0.1,\, 0.9]$, but limit our focus to $\eta \in \{0.2, 0.3\}$; the reason is that for smaller values of $\eta$, the point clouds remain overly fragmented across most layers, obscuring meaningful trends in connectivity; conversely, for larger values of $\eta$, the representations become rapidly connected, yielding trivial topological summaries. The chosen range thus balances sensitivity to topological changes while enabling consistent comparison across layers and architectures.

Because the MNIST analysis includes only a small number of representative architectures, correlation-based analyses between simplification timing (COM) and model size are statistically underpowered. Therefore, we adopt a distributional comparison approach, visualizing the COM values across models using violin plots (\figureref{fig:com_comparison_mnist}). Despite the lack of a known ground-truth topology for MNIST, we observe trends consistent with both our synthetic experiments and prior studies of feedforward networks \citep{Naitzat2020}. In particular, PCNs exhibit progressive topological simplification across layers, as reflected by a decrease in $\beta_0$, with the onset and rate of simplification influenced by architectural constraints and activation functions.



We do not report MNIST reconstructions. PCN reconstructions on MNIST are known to be difficult \citep{orchard2019genpcn}, and in our runs the reconstructions were unstable or low quality. We therefore restrict the MNIST study to representational topology.

\begin{figure}[htbp]
\floatconts
    {fig:com_comparison_mnist}
    {
        \caption{Violin plots of $\mathrm{COM}(\beta_0 + \beta_1)$ for three MNIST architectures ($\tanh$ activation), ordered left to right by decreasing total hidden-layer width, computed from $30$ independently trained networks per configuration. Lower COM indicates earlier topological simplification. For meaning of architecture names, refer to \tableref{tab:mnist_models}. Model sizes are $4096$, $1920$, and $1440$, respectively.}
        \vspace{-2\baselineskip}
    }
    {%
        \includegraphics[width=0.47\linewidth]{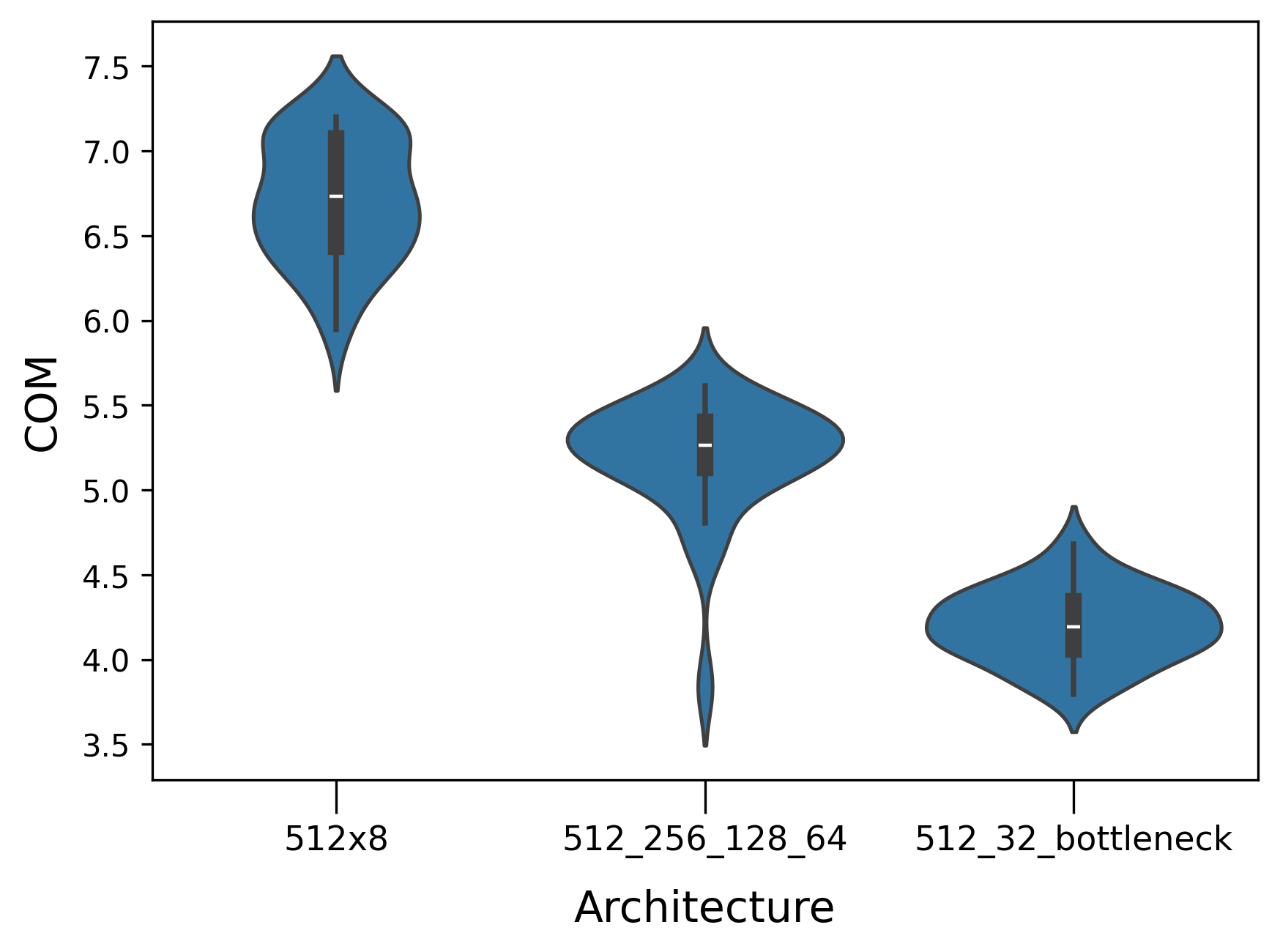}
        \vspace{-1.75\baselineskip}
    }%
\end{figure}

\subsection{Comparison with Feedforward Networks}
\label{sec:ffn_comparison}

We compare PCNs trained on the synthetic dataset with matched multilayer perceptrons (MLPs) trained to the same accuracy. For each architecture, activation function, and trained seed/model, we compute a single $\mathrm{COM}(\beta_0)$ value from the persistence diagrams. We then bootstrap seed-level means with replacement and compare $\Delta_{\mathrm{COM}}=\overline{\mathrm{COM}}_{\mathrm{PCN}}-\overline{\mathrm{COM}}_{\mathrm{MLP}}$.

Our results show a consistent positive COM gap: PCNs have larger COMs than matched MLPs for all architectures and activations tested. The bootstrap analysis yielded a p-value of $p < 10^{-4}$, rejecting the null hypothesis that $\Delta_\mathrm{COM} \leq 0$. The average COM difference $\Delta_\mathrm{COM}$ was $3.6$, meaning our PCNs simplified topology on average $3.6$ layers later than an MLP of the same architecture and activation. See \figureref{fig:bootstrap_difference_distributions} for the complete results.

\section{Conclusion}
In this work, we use persistent homology to study how representation topology evolves across layers in predictive coding networks. Our results show that topological simplification is not uniform across architectures: smaller models simplify earlier, while larger models (trained to the same accuracy) preserve topological structure deeper into the network. We also find that earlier simplification is correlated with weaker inversion quality.  Models that simplify rapidly tend to discard structurally relevant information before it can be reconciled with top-down predictions, resulting in poorer generative reconstructions.

Together, these findings suggest a capacity--topology--reconstruction tradeoff in PCNs: increasing capacity can delay topological simplification, improving invertibility, whereas smaller models with earlier simplification produce more compressed but less reconstructible representations. More broadly, our results demonstrate that persistent homology can serve as a useful quantitative probe of learned representations in recurrent, bidirectional architectures. Future work can test whether the same trends hold across broader datasets and architectures, and can develop refined reconstruction-aware topological metrics (such as extensions of Mean Reconstruction Distance) that better capture semantic fidelity in real-world data.

\bibliography{main}

\clearpage
\appendix

\setcounter{figure}{0}
\renewcommand{\thefigure}{A.\arabic{figure}}
\renewcommand{\theHfigure}{A.\arabic{figure}}
\setcounter{equation}{0}
\renewcommand{\theequation}{A.\arabic{equation}}
\setcounter{table}{0}
\renewcommand{\thetable}{A.\arabic{table}}

\section{Predictive Coding Network Formulation}
\label{sec:PCN_appendix}

We implement our models using the \texttt{JAX}-based \texttt{PCX} library \citep{pinchetti2025pcx}, which follows the formulation of PCNs introduced by
\citet{whittington2017approximation}.

Each layer $\ell$ contains neural activities $x_i^{(\ell)}$, where
\mbox{$i \in \{1,\dots,n_\ell\}$} indexes units within the layer. Top-down predictions are defined via a generative model.
The predicted mean activity $\mu_i^{(\ell)}$ at layer $\ell$ is given by
\begin{equation}
\mu_i^{(\ell)}
=
\sum_{j=1}^{n_{\ell+1}} \theta_{i,j}^{(\ell+1)}
\, f\!\left(x_j^{(\ell+1)}\right),
\label{eq:pcn_mu}
\end{equation}
where $\theta_{i,j}^{(\ell+1)}$ denotes the synaptic weight from unit $j$
in layer $\ell+1$ to unit $i$ in layer $\ell$, and $f(\cdot)$ is a (possibly nonlinear)
activation function.

Prediction errors are defined as variance-normalized residuals
\begin{equation}
\epsilon_i^{(\ell)}
=
\frac{x_i^{(\ell)} - \mu_i^{(\ell)}}{\Sigma_i^{(\ell)}},
\label{eq:pcn_error}
\end{equation}
where $\Sigma_i^{(\ell)}$ denotes the variance associated with unit $i$
at layer $\ell$.

Under a Gaussian noise assumption, the variational free energy is given by
\begin{equation}
F
=
\frac{1}{2}
\sum_{\ell=0}^L
\sum_{i=1}^{n_{\ell}}
\frac{\left(x_i^{(\ell)} - \mu_i^{(\ell)}\right)^2}{\Sigma_i^{(\ell)}},
\label{eq:pcn_energy}
\end{equation}
where the $L$\textsuperscript{th} layer is the last hidden layer.

Inference corresponds to gradient descent on this free energy with respect to the neural activities. The resulting continuous-time dynamics for unit $b$ in layer $a$ are
\begin{equation}
\dot{x}_b^{(a)}
=
-\epsilon_b^{(a)}
+
\sum_{i=1}^{n_{a-1}}
\epsilon_i^{(a-1)}
\, \theta_{i,b}^{(a)}
\, f'\!\left(x_b^{(a)}\right),
\label{eq:pcn_inference}
\end{equation}
where $f'(\cdot)$ denotes the derivative of the activation function.

Learning proceeds via gradient descent on the variational free energy
with respect to the synaptic parameters.
The resulting gradient is given by \citet[Eq.~(2.21)]{whittington2017approximation}:
\begin{equation}
\frac{\partial F}{\partial \theta_{i,j}^{(\ell)}}
=
\epsilon_i^{(\ell-1)}\, f\!\left(x_j^{(\ell)}\right).
\end{equation}
This confirms that both inference and learning depend only on locally available pre- and post-synaptic signals.

Because inference is defined by minimizing a shared energy functional rather than propagating signals through a fixed forward pathway, PCNs naturally admit approximate inversion by clamping higher-level states and inferring lower-level representations through the same dynamics.

\subsection{Inversion}
\label{sec:inversion_appendix}
The recurrent, bidirectional dynamics of PCNs naturally enable what we refer to as \emph{inversion}: the generation of input-level representations from fixed higher-level states \citep{rao1999predictive, sun2020inversion}. In our implementation, inversion is performed by clamping the output-layer target and then directly optimizing the input through a top-down reconstruction of the lower layers that best explains that imposed state under the learned generative parameters. Concretely, we solve an energy-minimization problem of the form
\begin{equation}
    \hat{z} \in \underset{z}{\mathrm{argmin}}\ F(z, y_\text{target}),
\end{equation}
where $y_\text{target}$ is a one-hot class vector and $F$ is the variational free energy (\equationref{eq:pcn_energy}).

To keep reconstructions bounded, we optimize this unconstrained auxiliary variable $\hat{z}$ and set $\hat{x} = \tanh(\hat{z})$ to be our ``reconstruction'', ensuring $\hat{x} \in (-1, 1)^d$; this matches the natural domain of our datasets. Starting from a random initialization of $z$, we perform a fixed number of gradient descent steps on the energy function $F$. This stochasticity allows the same target output to yield a distribution of plausible reconstructions rather than a single deterministic result.

We leverage this inversion mechanism in \sectionref{sec:mrd} to assess the internal consistency and representational fidelity of the learned model; see \figureref{fig:recons_comparison} for a visualization.

\begin{figure}[htbp]
\floatconts
    {fig:recons_comparison}
    {\caption{Reconstructed inputs from two PCNs (red), overlaid on the true manifold $M_a$ (green). Reconstructions by \texttt{30x8\_leaky} (left) remain localized within components, while those by \texttt{30x4\_10x4\_relu} (right) are substantially more scattered and disconnected, indicating weaker approximate invertibility, aligning with the former's significantly lower MRD ($0.024$ vs. $0.122$, respectively).}}
    {%
        \subfigure[\texttt{30x8\_leaky} model] {%
            \label{fig:30x8_leaky_recons}
            \includegraphics[width=0.45\linewidth]{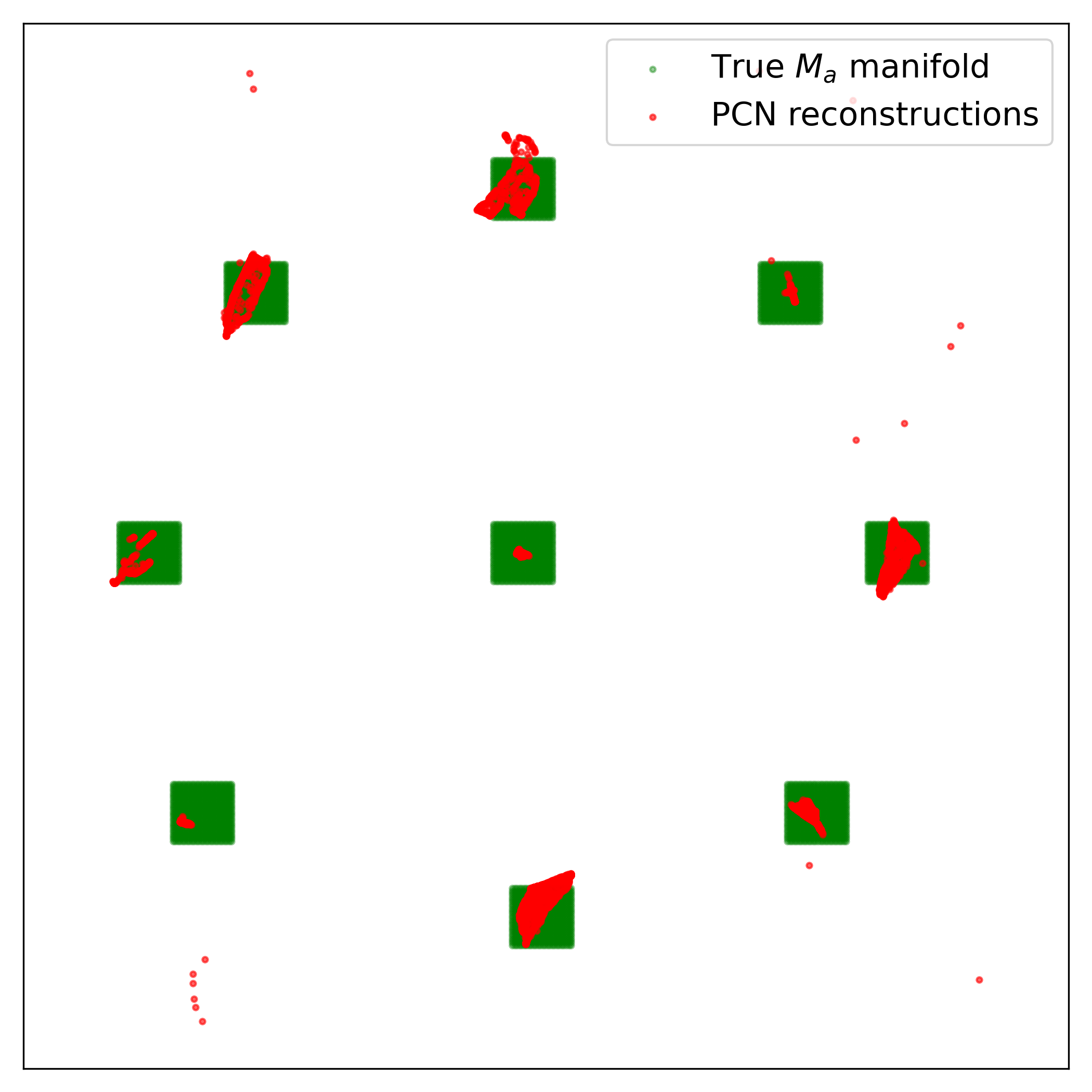}
        }
        \ \ 
        \subfigure[\texttt{30x4\_10x4\_relu} model] {%
            \label{fig:30x4_10x4_relu_recons}
            \includegraphics[width=0.45\linewidth]{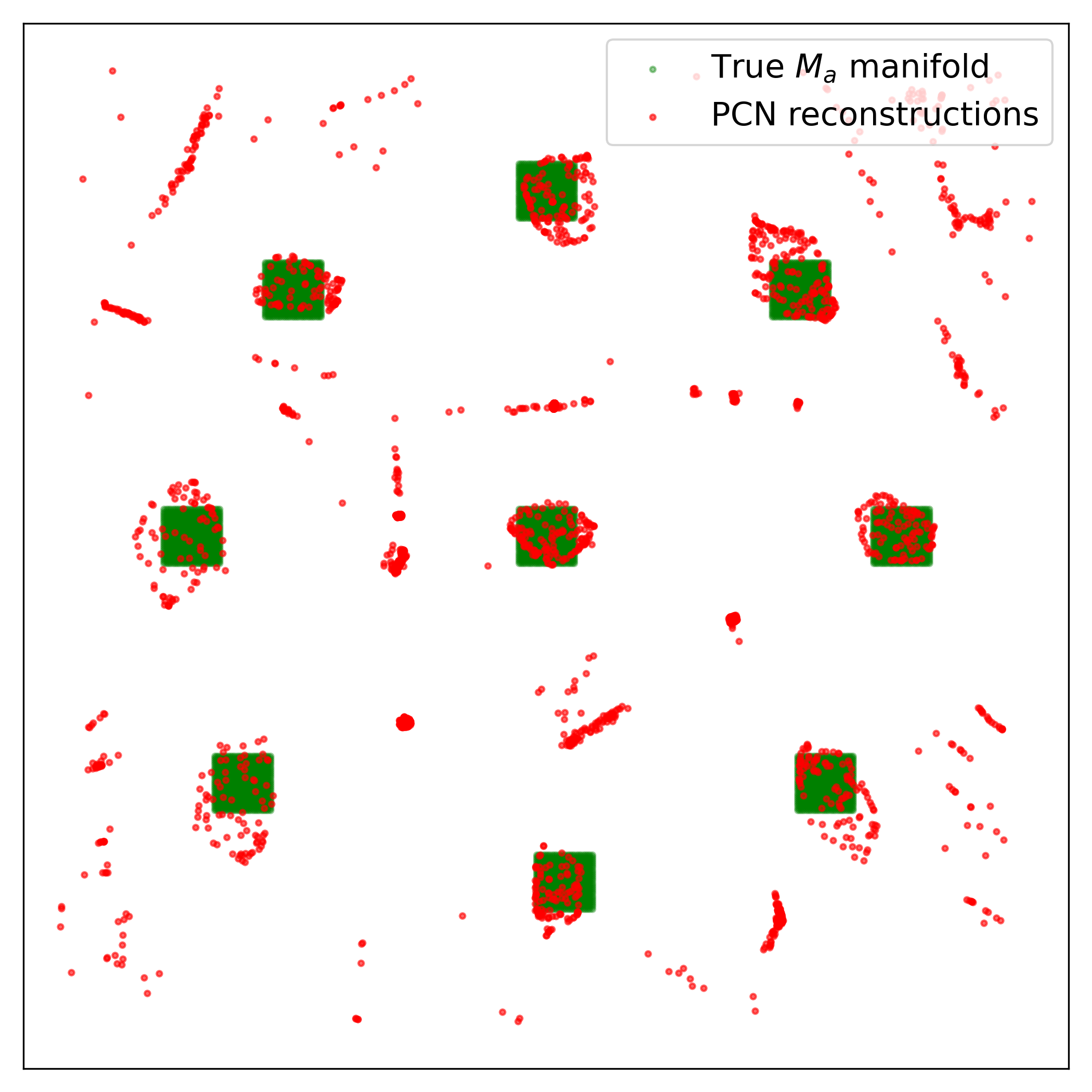}
        }
        \vspace{-\baselineskip}
    }
\end{figure}

\section{Model Details}
\label{app:model_details}

\setcounter{figure}{0}
\renewcommand{\thefigure}{B.\arabic{figure}}
\renewcommand{\theHfigure}{B.\arabic{figure}}
\setcounter{equation}{0}
\renewcommand{\theequation}{B.\arabic{equation}}
\setcounter{table}{0}
\renewcommand{\thetable}{B.\arabic{table}}

Below is a summary of the models used in our experiments, including activation functions used and the number of neurons in each layer. Our implementation of Leaky ReLU weighs negative inputs by $0.2$.

\begin{table}[ht]
\floatconts
    {tab:all_models}
    {\caption{Summary of architectures trained on the synthetic dataset with their hidden layer widths and activation functions used.}}
    {%
        \vspace{-\baselineskip}
        \begin{tabular}{|l|l|l|}
        \hline
        \textbf{Name} & \textbf{Activations} & \textbf{Neurons in Each Layer} \\ \hline
        
        30x8 & ReLU, $\tanh$, Leaky ReLU & 30, 30, 30, 30, 30, 30, 30, 30 \\ \hline
        24x8 & ReLU, $\tanh$, Leaky ReLU & 24, 24, 24, 24, 24, 24, 24, 24 \\ \hline
        18x8 & ReLU, $\tanh$, Leaky ReLU & 18, 18, 18, 18, 18, 18, 18, 18 \\ \hline
        30x4\_24x4 & ReLU, $\tanh$, Leaky ReLU & 30, 30, 30, 30, 24, 24, 24, 24 \\ \hline
        30x4\_18x4 & ReLU, $\tanh$, Leaky ReLU & 30, 30, 30, 30, 18, 18, 18, 18 \\ \hline
        30x4\_12x4 & ReLU, $\tanh$, Leaky ReLU & 30, 30, 30, 30, 12, 12, 12, 12 \\ \hline
        30x4\_10x4 & ReLU, $\tanh$, Leaky ReLU & 30, 30, 30, 30, 10, 10, 10, 10 \\ \hline
        30\_28\_26\_24\_22\_20\_18\_16 & ReLU, $\tanh$, Leaky ReLU & 30, 28, 26, 24, 22, 20, 18, 16 \\ \hline
        30\_25\_20\_15\_10x4 & ReLU, $\tanh$, Leaky ReLU & 30, 25, 20, 15, 10, 10, 10, 10 \\ \hline
        30\_15\_30x6 & ReLU, $\tanh$, Leaky ReLU & 30, 15, 30, 30, 30, 30, 30, 30 \\ \hline
        Alternating (30, 15) & ReLU & 30, 15, 30, 15, 30, 15, 30, 15 \\ \hline
        15x4\_30x4 & ReLU, $\tanh$ & 15, 15, 15, 15, 30, 30, 30, 30 \\ \hline
        
        \end{tabular}
    }
\end{table}

\begin{table}[htbp]
\floatconts
    {tab:mnist_models}
    {\caption{Summary of architectures trained on MNIST with their hidden layer widths and activation functions used.}}
    {%
        \vspace{-\baselineskip}
        \begin{tabular}{|l|l|l|}
        \hline
        \textbf{Name} & \textbf{Activations} & \textbf{Neurons in Each Layer} \\ \hline
        
        512x8 & ReLU, $\tanh$, Leaky ReLU, Softplus & 512, 512, 512, 512, 512, 512, 512, 512 \\ \hline
        256x8 & ReLU, $\tanh$, Leaky ReLU, Softplus & 256, 256, 256, 256, 256, 256, 256, 256 \\ \hline
        512\_32 & ReLU, $\tanh$, Leaky ReLU, Softplus & 512, 384, 256, 192, 128, 96, 64, 32 \\ \hline
        512\_256\_128\_64  & ReLU, $\tanh$, Leaky ReLU & 512, 512, 256, 256, 128, 128, 64, 64 \\ \hline
        512\_32\_bottleneck & ReLU, $\tanh$ & 512, 256, 128, 64, 32, 64, 128, 256 \\ \hline
        
        \end{tabular}
    }
\end{table}
\newpage
\section{Topology Tracking Procedure}
\label{sec:topology_tracking_appendix}

\setcounter{figure}{0}
\renewcommand{\thefigure}{C.\arabic{figure}}
\renewcommand{\theHfigure}{C.\arabic{figure}}
\setcounter{equation}{0}
\renewcommand{\theequation}{C.\arabic{equation}}
\setcounter{table}{0}
\renewcommand{\thetable}{C.\arabic{table}}

\subsection{Synthetic Data}
\label{sec:topology_tracking_d1_appendix}

We adopt and extend the topological analysis framework of \citet{Naitzat2020} to study how the topology of a point-cloud dataset evolves as it propagates through the hidden layers of a PCN. All models are first trained to convergence, achieving at least $99.9\%$ test accuracy. 

As in \citet{Naitzat2020}, we compute and track persistent homology on only one of the two manifolds in the synthetic dataset, denoted $M_a$ in \sectionref{sec:datasets} and \figureref{fig:D1_picture}. Concretely, before extracting layer-wise representations, we filter the dataset to retain only points belonging to $M_a$ (i.e., a single class) and then compute Betti numbers on this restricted point cloud. This restriction is necessary because the full dataset is a union of two manifolds; computing homology on the union would conflate the topology of manifold $M_a$ with that of manifold $M_b$ and with inter-manifold separations, obscuring the within-manifold simplification behavior we aim to measure.

To ensure comparability across layers with varying activation scales, we compute topology using the unweighted geodesic distance on a $k$-NN graph constructed over each layer’s representations. By assigning unit weight to all $k$-NN edges, this metric normalizes distances within each latent space, reducing sensitivity to absolute scale and enabling the use of a fixed persistent homology threshold $\eta = 2.5$ across layers.

The choice of the $k$-NN parameter $k$ is critical: a $k$ too small results in a disconnected graph ($\beta_0 > \text{ground truth}$), while a $k$ too large induces ``short-circuiting'' that collapses the manifold's intrinsic structure. To determine an optimal $k$ for the synthetic dataset, we perform a Monte Carlo cross-validation procedure:
\begin{enumerate}
    \item Subsampling: In each of $n=1000$ trials, we randomly sample a fraction ($25\%$) of the synthetic manifold to simulate varying densities.
    \item Iterative Search: For each subset, we iteratively increase $k$ from $1$ to $100$, constructing the corresponding geodesic distance matrix.
    \item Homological Matching: We compute the zeroth Betti number ($\beta_0$) using the distance matrix and identify the minimal $k$ required to recover the ground-truth topology of manifold $M_a$ ($\beta_0 = 9$).
\end{enumerate}
The distribution converges to a mode of $k=14$. This value consistently recovers the true connectivity of the manifold while remaining below the threshold of over-connectivity. To maintain consistency across all comparative experiments, we fix $k=14$ and utilize a single representative $25\%$ subset whose topology exactly matches the ground-truth manifold signature. This aligns with the methodology used in \cite{Naitzat2020}.

Persistent homology is then computed via Vietoris–Rips complexes using the \texttt{Ripser} Python package \citep{bauer2021ripser}, itself relying on foundational work from \citet{zomorodian2004ph} and \citet{edelsbrunner2002persistence}. A crucial result comes from \citet{attali2013vrcomplex}, showing that for a large enough sample size and at a sufficiently small scale, the topology of the VR-complex of a manifold reflects the true topology of that manifold.

\subsection{Real-World Data}
\label{sec:topology_tracking_mnist_appendix}

For real-world datasets such as MNIST, the true topology of the underlying data manifold is unknown. Consequently, unlike in synthetic settings, there is no way to select or validate a $k$-NN construction by verifying recovery of the correct input-layer topology, motivating the use of other distance metrics.

The most immediate alternative is Euclidean distance; however, when computing persistent homology on layer-wise activations, raw Euclidean distances are not directly comparable across layers due to arbitrary expansions or contractions induced by network weights \citep{Naitzat2020}. Without normalization, changes in persistent homology may therefore reflect trivial scaling effects rather than meaningful changes in representational structure.

To address this issue, we follow \citet{wheeler2021}, who adopt a simple per-layer metric normalization procedure: For a fixed layer $\ell$, let $X^{(\ell)} = \{x_1^{(\ell)}, \dots, x_n^{(\ell)}\} \subset \mathbb{R}^{n_\ell}$ denote the activations of $n$ distinct data points at that layer. We compute the full pairwise Euclidean distance matrix and rescale it by its maximum entry, defining a normalized metric
\begin{equation}
\label{eq:distance_l}
d_\ell(x_i, x_j)
\;=\;
\frac{\|x_i^{(\ell)} - x_j^{(\ell)}\|_2}
{\max_{p,q} \|x_p^{(\ell)} - x_q^{(\ell)}\|_2}.
\end{equation}
This normalization ensures that the diameter of the metric space $(X^{(\ell)}, d_\ell)$ is equal to one for every layer, thereby placing all layers on a common distance scale.

\cite{edelsbrunner2010computational} show that this per-layer distance normalization does not alter the intrinsic topology recovered at any fixed layer, but merely rescales the filtration parameter, allowing Betti numbers to be compared across layers without confounding effects due to varying magnitudes of network weights.

Persistent homology is otherwise calculated identically as in \appendixref{sec:topology_tracking_d1_appendix}, with the digit-$0$ class being tracked. For computational efficiency, we restrict attention to homology dimensions $k \in \{0,1\}$, which capture connected components and one-dimensional loops, respectively.

\newpage

\section{Additional Figures}
\setcounter{figure}{0}
\renewcommand{\thefigure}{D.\arabic{figure}}
\renewcommand{\theHfigure}{D.\arabic{figure}}
\setcounter{equation}{0}
\renewcommand{\theequation}{D.\arabic{equation}}
\setcounter{table}{0}
\renewcommand{\thetable}{D.\arabic{table}}


\begin{figure}[htbp]
\floatconts
    {fig:bootstrap_difference_distributions}
    {\caption{Bootstrap distributions of COM mean differences between matched PCNs and MLPs. Each panel shows $\Delta_{\mathrm{COM}}=\mathrm{COM}_{\mathrm{PCN}}-\mathrm{COM}_{\mathrm{MLP}}$ for one architecture--activation pair. The dashed vertical line marks zero; all distributions lie to the right of zero, indicating larger COM for PCNs.}}
    {
        \includegraphics[width=0.95\linewidth]{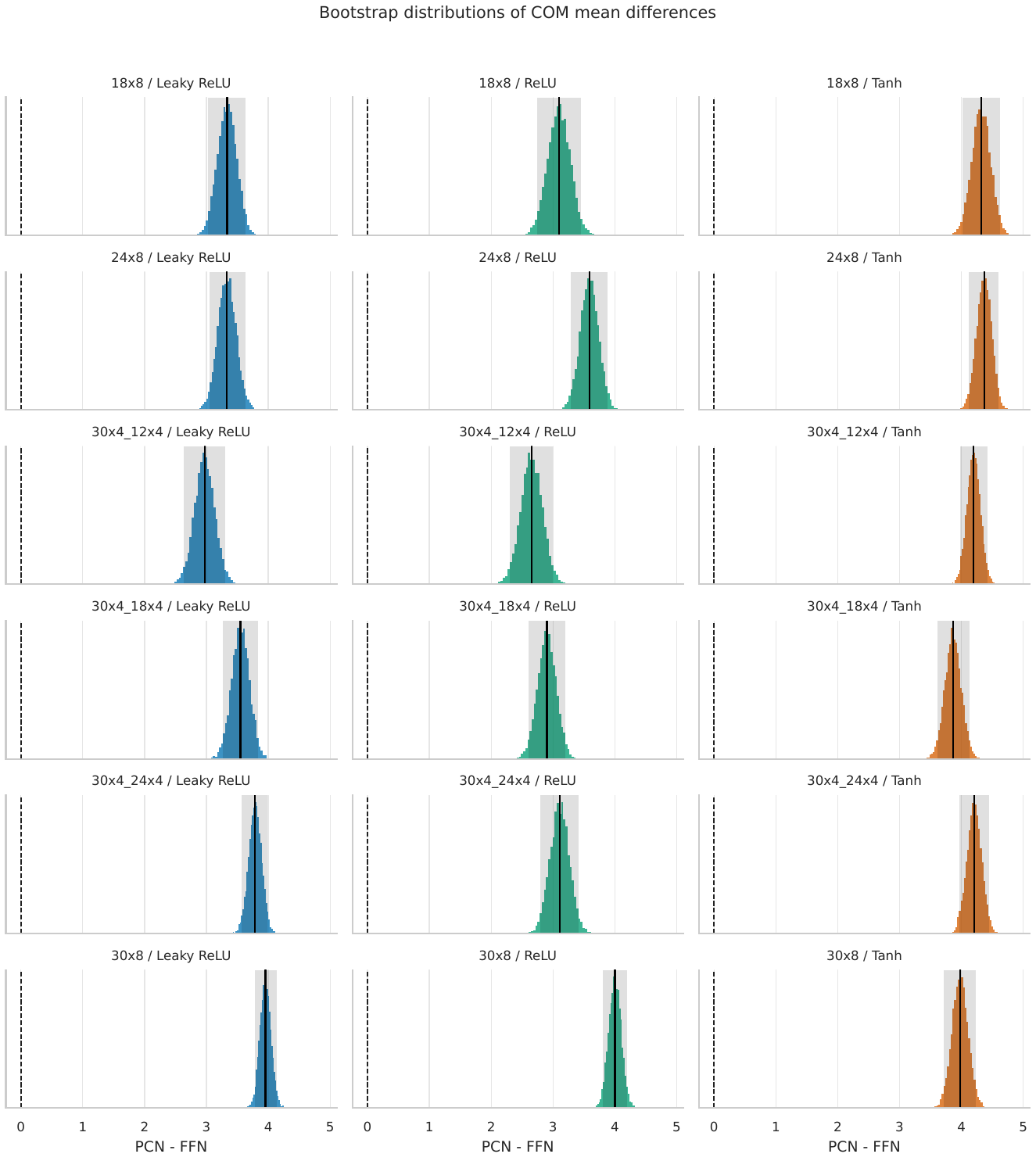}
        \vspace{-\baselineskip}
    }
\end{figure}

\end{document}